\documentclass[10pt,twocolumn,letterpaper]{article}

\usepackage{3dv}
\usepackage{times}
\usepackage{epsfig}
\usepackage{graphicx}
\usepackage{amsmath}
\usepackage{amssymb}
\usepackage{subcaption}
\usepackage{algorithm}
\usepackage{graphicx,booktabs,array}

\newcolumntype{C}{>{\centering\arraybackslash}m{10em}}

\pdfoutput=1

\threedvfinalcopy 

\def\threedvPaperID{****} 

\ifthreedvfinal\fi
\begin{document}

\title{Bilateral Operators for Functional Maps}
\author{Gautam Pai \hspace{1cm} Mor Joseph-Rivlin \hspace{1cm} Ron Kimmel\\
Technion - Israel Institute Of Technology\\
{\tt\small \{paigautam,ron\}@cs.technion.ac.il, mor1joseph@campus.technion.ac.il}}

\maketitle

\begin{abstract}
A majority of shape correspondence frameworks are based on devising pointwise and pairwise constraints on the correspondence map. The functional maps framework allows for formulating these constraints in the spectral domain. 
In this paper, we develop a functional map framework for the shape correspondence problem by constructing pairwise constraints using point-wise descriptors. 
Our core observation is that, every point-wise descriptor allows for the construction a pairwise kernel operator whose low frequency eigenfunctions depict regions of similar descriptor values at various scales of frequency. 
By aggregating the pairwise information from the descriptor and the intrinsic geometry of the surface encoded in the heat kernel, we construct a hybrid kernel and call it the bilateral operator. 
Analogous to the edge preserving bilateral filter in image processing, the action of the bilateral operator on a function defined over the manifold yields a descriptor dependent local smoothing of that function. 
By forcing the correspondence map to commute with the Bilateral operator, we show that we can maximally exploit the information from a given set of pointwise descriptors in a functional map framework. 
\end{abstract}

\section{Introduction}
Shape correspondence between a pair of surfaces is the task of estimating a semantically meaningful map from one surface onto the other. In geometry processing, surfaces are typically represented as compact two-dimensional Riemannian manifolds. Given two such surfaces $\mathcal{X}$ and $\mathcal{Y}$, the correspondence map $\varphi : \mathcal{X} \rightarrow \mathcal{Y}$ is typically obtained as the solution to an optimization problem where the notion of a good or bad correspondence is explicitly modeled by the objective function. 

Typically, such cost functions comprise of a combination of two components: \emph{pointwise descriptor preservation} or function preservation constraints, and \emph{pairwise descriptor preservation} or operator commutativity constraints. A point-wise descriptor associates a scalar value to every point on the surface. Therefore, any candidate for its corresponding point on the second shape must have an equivalent descriptor value. For a surface discretized by $n$ points, a single point-wise descriptor is a vector of size $n \times 1$. Notable examples are the heat kernel signature (HKS) \cite{sun2009concise}, the wave kernel (WKS) \cite{aubry2011wave}, shape context signatures \cite{kokkinos2012intrinsic} etc. 

In contrast, a pairwise descriptor associates a scalar value for every \emph{pair} of points on each shape. Therefore, given a pair of points on the first shape, a correspondence map must result in choosing corresponding pairs of points on the second shape yielding equivalent pairwise descriptor values. For an $n$ point surface, a single pairwise descriptor is a matrix of size $n \times n$. Typical examples include, inter-geodesic distance matrices \cite{bronstein2006generalized} and heat kernels \cite{ovsjanikov2010one,vestner2017efficient} of the surfaces. Pairwise descriptors can also be interpreted as operators acting on functions defined over the surface. Therefore the constraint of preserving a pairwise descriptor is equivalent to the argument that applying the operator to any function and transferring it by correspondence must be equivalent to transferring the function first and then operating on it, thus commuting with the correspondence map.  

The functional maps paradigm \cite{ovsjanikov2012functional}, provides for an elegant linear algebraic method of estimating correspondence maps between surfaces. Under the assumption of approximate isometry, the main idea is to formulate the pairwise and pointwise constraints in the spectral domain of the shapes and represent the map using the eigenfunctions of the laplace beltrami operators of the surfaces. This formulation enables replacing a combinatorial non-convex optimization problem of finding permutations with a simple linear least squares estimation yielding a linear map between the functional spaces of the pair of manifolds. 

In most practical applications, it is usually convenient to find reliable pointwise descriptors like point or part correspondences. Eg: shape segments \cite{litman2012stable}, spectral descriptors \cite{sun2009concise,aubry2011wave,litman2013learning}, histograms of local features \cite{tombari2010unique} etc. that are relatively easier to identify and compute. The authors of \cite{nogneng2017informative} demonstrate that the information from a point-wise descriptor can be extended to construct a linear operator from it. They propose an operator (or pairwise descriptor) that acts on a function through a point-wise multiplication of the descriptor. Imposing a commutativity constraint with this operator leads to an improved functional map estimation from the same set of known corresponding point-wise descriptors.

Advancing this idea of constructing a pairwise operator from a pointwise descriptor, we explore a novel operator which we call the bilateral operator. The bilateral operator is a smoothing operator, where the weights of averaging are obtained by a combination of intrinsic and descriptor similarities. Each pointwise descriptor can be used to define a pairwise bilateral operator whose action indicates descriptor dependent smoothing of a function over the manifold. By requiring that the correspondence map commute with such an operator, we show a considerable increase in correspondence quality in comparison to previous functional map approaches. The intuition behind our method is that combination of geodesic and descriptor closeness yields high quality correspondence. 
This means that pairs of points on one shape that are close geodesically or have similar descriptor values correspond very distinctly to pairs of such similar points on the other shape, given the availability of good quality corresponding pointwise descriptors. 


\section{Background}
Without loss of generality, let us consider a pair of shapes $\mathcal{X}$ and $\mathcal{Y}$ that are sampled consistently by $n$ points each. We consider the discrete counterpart of the correspondence map $\varphi : \mathcal{X} \rightarrow \mathcal{Y}$ and observe that it can be represented as a permutation matrix $\boldsymbol{\Pi} \in \{0,1\}^{n \times n}$ satisfying 
$\mathbf{\Pi}^{\top} \mathbf{1}=\mathbf{\Pi} \mathbf{1}=\mathbf{1}$. where $\mathbf{1}$ is a column vector of ones.  A majority of the shape correspondence algorithms can be expressed as an energy minimization problem having the following form \cite{vestner2017efficient}:
\begin{equation}
\boldsymbol{\Pi}^{*} = \arg \min _{\mathbf{\Pi} \in \mathcal{P}_{n}} h(\boldsymbol{\Pi}) + \alpha g(\boldsymbol{\Pi})
\label{eq:energy_min}
\end{equation}
where $\mathcal{P}_{n}$ denotes the space of $n \times n$ permutation matrices. 

Given a set of corresponding point-wise descriptors: $\{ f_{\mathcal{X}}^{(i)}, f_{\mathcal{Y}}^{(i)} \in  \mathbb{R}^{n\times 1}\}_{i=1}^{i=N}$, the pointwise descriptor preservation term can be written as:
\begin{equation}
h(\boldsymbol{\Pi}) = \sum_{i=1}^{i=N} ||\boldsymbol{\Pi} f_{\mathcal{X}}^{(i)} - f_{\mathcal{Y}}^{(i)} ||^2
\label{pointwise}
\end{equation}
and similarly, for a set of corresponding pairwise descriptors $\{ O_{\mathcal{X}}^{(i)}, O_{\mathcal{Y}}^{(i)} \in  \mathbb{R}^{n\times n}\}_{i=1}^{i=M}$, the pairwise descriptor preservation (or operator commutativity constraint) is given by:
\begin{equation}
g(\boldsymbol{\Pi}) = \sum_{i=1}^{i=M} ||\boldsymbol{\Pi} O_{\mathcal{X}}^{(i)} - O_{\mathcal{Y}}^{(i)} \boldsymbol{\Pi}||^2
\label{pairwise}
\end{equation}

The functional maps paradigm \cite{ovsjanikov2012functional} allows for solving this problem efficiently by leveraging the knowledge of a compact basis for representing functions on each surface. Typically, the eigenfunctions of the Laplace Beltrami operator of each surface are used, since they form a convenient basis to represent functions and more importantly, they are proven to be optimal for representing smooth functions over manifolds \cite{aflalo2015optimality}. 
\begin{figure*}[t]
\centering
\begin{subfigure}[h]{0.28\textwidth}
\includegraphics[scale=0.25]{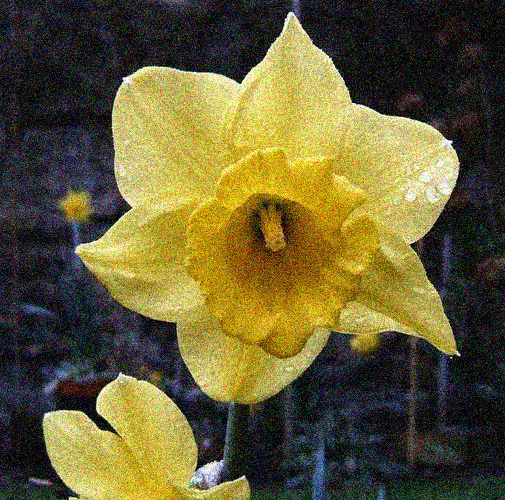}
\caption{}
\end{subfigure}
\begin{subfigure}[h]{0.28\textwidth}
\includegraphics[scale=0.25]{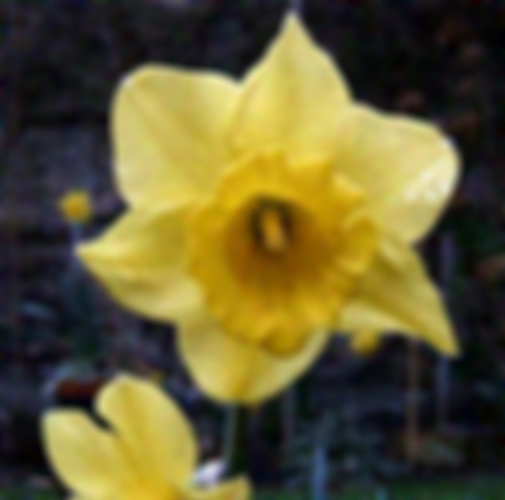}
\caption{}
\end{subfigure}
\begin{subfigure}[h]{0.28\textwidth}
\includegraphics[scale=0.25]{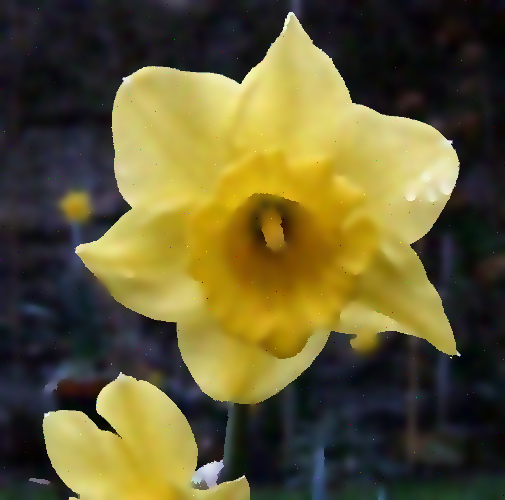}
\caption{}
\end{subfigure}
{\caption{ {\bf Image Filtering}: (a) A noisy image (b) Image after standard gaussian smoothing of Equation \ref{eq:gauss_smooth}. (c) Image after Bilateral filtering in \ref{eq:bilateral_smooth}. In contrast to the result of the gaussian smoothing, the bilteral filter is shown to distinctly preserve dominant edges in the image, enabled by the additional image dependent kernel.}
\label{bilateral_image}}
\end{figure*} 

To fix notations, let $\mathcal{X}$ be a two-dimensional Riemannian manifold. Consider the space of functions defined on the manifold: $ L^{2}(\mathcal{X}) = \left\{f : \mathcal{X} \rightarrow \mathbb{R} |\langle f, f\rangle_{\mathcal{X}}<\infty\right\}$.
The positive semi-definite Laplace-Beltrami operator (LBO) $\Delta_{\mathcal{X}}$ generalizes the notion of the Laplacian operator from Euclidean spaces to surfaces. $\Delta_{\mathcal{X}}$ admits an orthonormal eigendecomposition $\Delta_{\mathcal{X}} \phi_{i}=\lambda_{i} \phi_{i}$, where the eigenvalues form a discrete spectrum $0=\lambda_{1} \leq \lambda_{2} \leq \ldots$ and the eigenfunctions $\Phi =  \{ \phi_{1}, \phi_{2}, \dots$ \} form an orthonormal basis for $ L^{2}(\mathcal{X})$, thereby allowing us to expand any smooth function  $f \in L^{2}(\mathcal{X})$ as a Fourier series 
\begin{equation} 
f(x)=\sum_{i \geq 1}\left\langle\phi_{i},f\right\rangle_{\mathcal{X}} \phi_{i}(x)
\end{equation}

Let $0=\mu_{1} \leq \mu_{2} \leq \ldots$ and $\Psi = \{ \psi_{1}, \psi_{2}, \dots \}$ be the corresponding eigenvalues and eigenfunctions of the LBO $\Delta_{\mathcal{Y}}$ on $\mathcal{Y}$. Then the functional representation of the correspondence map in the spectral basis is given by
\begin{equation}
    \boldsymbol{\Pi} = \Psi C \Phi^T
    \label{eq:fmap_spectral}
\end{equation}
Where the matrix C is a functional maps matrix, whose elements encode the coefficients of the correspondence map represented in the spectral domain. The main utility of the functional maps framework is that, a much larger $n \times n$ permutation can be approximated by a smaller $k \times k$ matrix C by using a few low frequency eigenfunctions $\Phi,\Psi \in \mathbb{R}^{n \times k}$ of the laplace beltrami operators $\Delta_{\mathcal{X}}$ and $\Delta_{\mathcal{Y}}$ respectively. 

By rewriting equation \ref{eq:energy_min} using the spectral approximation of \ref{eq:fmap_spectral}, the functional maps matrix can be obtained as the solution to the problem:
\begin{equation}
\begin{aligned}
C^{*} = \arg \min_{C} & \sum_{i=1}^{i=N} ||C \widehat{f}_{\mathcal{X}}^{(i)} - \widehat{f}_{\mathcal{Y}}^{(i)} ||^2 \\ +  \alpha & \sum_{i=1}^{i=M} ||C \widehat{O}_{\mathcal{X}}^{(i)} - \widehat{O}_{\mathcal{Y}}^{(i)} C||^2
\end{aligned}
\label{eq:fmap_energymin}
\end{equation}
where 
\begin{eqnarray}
\widehat{f}_{\mathcal{X}}^{(i)} &=& \Phi^T A_{\mathcal{X}}^T f_{\mathcal{X}}^{(i)}, \; 
\widehat{f}_{\mathcal{Y}}^{(i)} = \Psi^T A_{\mathcal{Y}}^T f_{\mathcal{Y}}^{(i)} \\
\widehat{O}_{\mathcal{X}}^{(i)} &=& \Phi^T A_{\mathcal{X}}^T O_{\mathcal{X}}^{(i)} A_{\mathcal{X}} \Phi \\
\widehat{O}_{\mathcal{Y}}^{(i)} &=& \Psi^T A_{\mathcal{Y}}^T O_{\mathcal{Y}}^{(i)} A_{\mathcal{Y}} \Psi
\end{eqnarray}
Where $A_{\mathcal{X}}$ and $A_{\mathcal{Y}}$ are the area matrices of the respective shapes. Different shape correspondence algorithms in literature can be viewed as particular cases of the loss functions \ref{eq:energy_min} or \ref{eq:fmap_energymin}, depending on whether the correspondence was solved in the spatial domain as a permutation or in the spectral domain as a functional map. Notable examples include \cite{nogneng2017informative}, which proposes a diagonal operator $ O_{\mathcal{X}}^{(i)} = \textrm{diag}(f_{\mathcal{X}}^{(i)}) \in \mathbb{R}^{n \times n}$ whos action is equivalent to a pointwise multiplication of the corresponding descriptors $f_{\mathcal{X}}^{(i)}$ and $f_{\mathcal{Y}}^{(i)}$. Similarly  \cite{dubrovina2011approximately,dubrovina2010matching,aflalo2016spectral} show the application of geodesic distances and \cite{vestner2017efficient} proposed the use of heat-kernels of the surfaces as pairwise descriptors. 
\begin{figure*}
\begin{center}
\hspace{-0.5cm}
\begin{subfigure}{0.2\textwidth}
\includegraphics[scale=0.1]{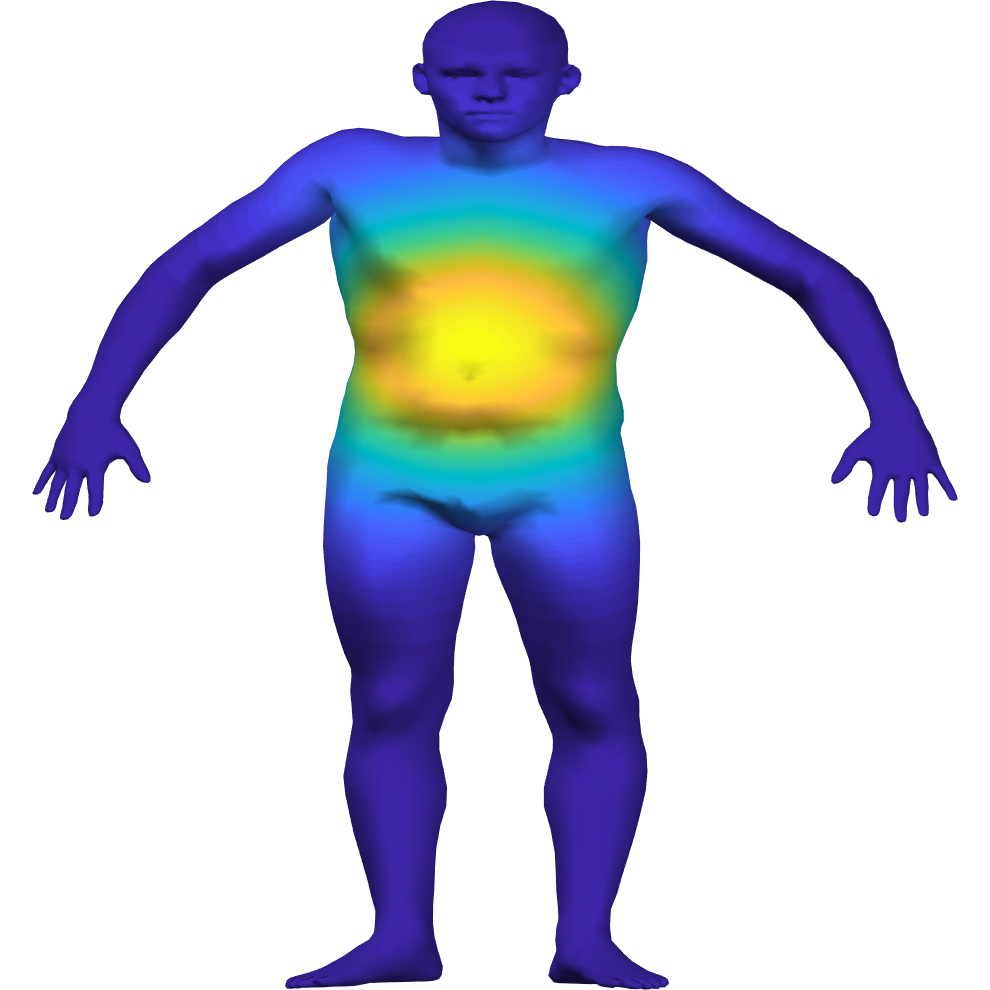}
\caption{}
\end{subfigure}
\begin{subfigure}{0.2\textwidth}
\includegraphics[scale=0.1]{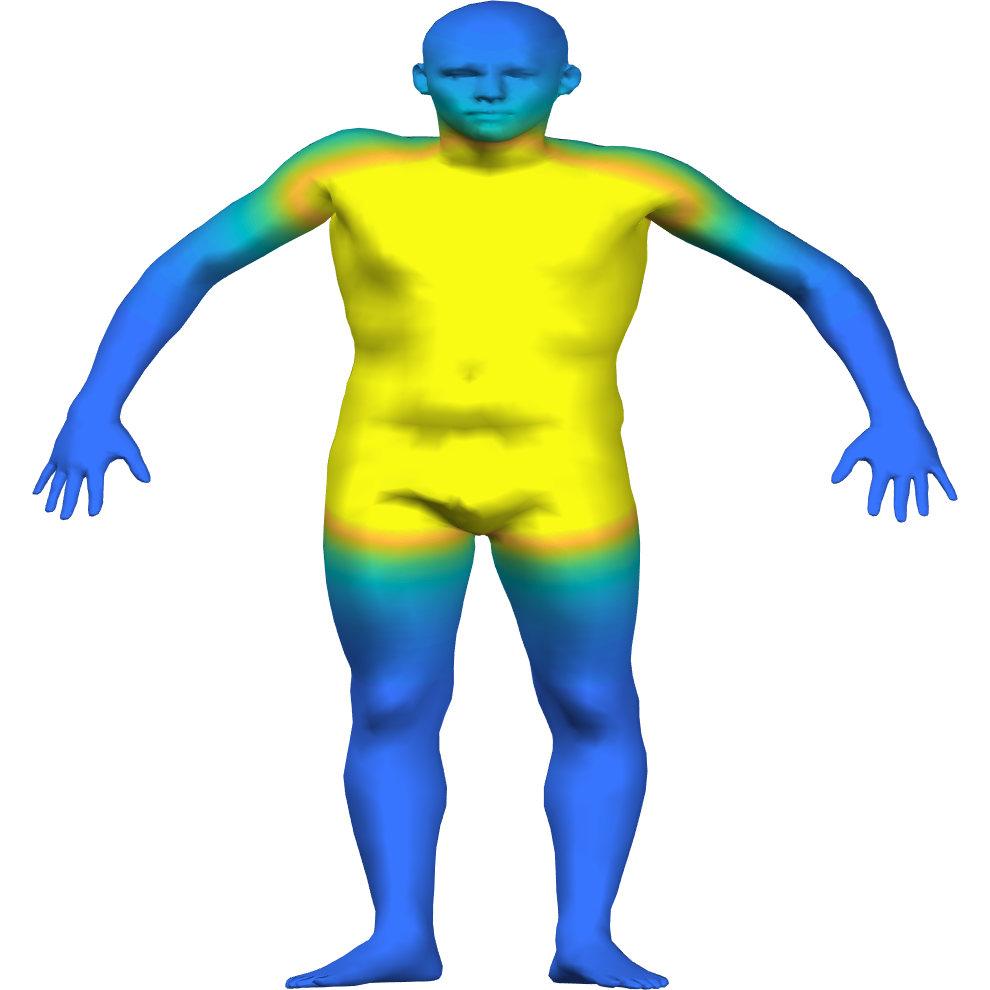}
\caption{}
\end{subfigure}
\begin{subfigure}{0.2\textwidth}
\includegraphics[scale=0.1]{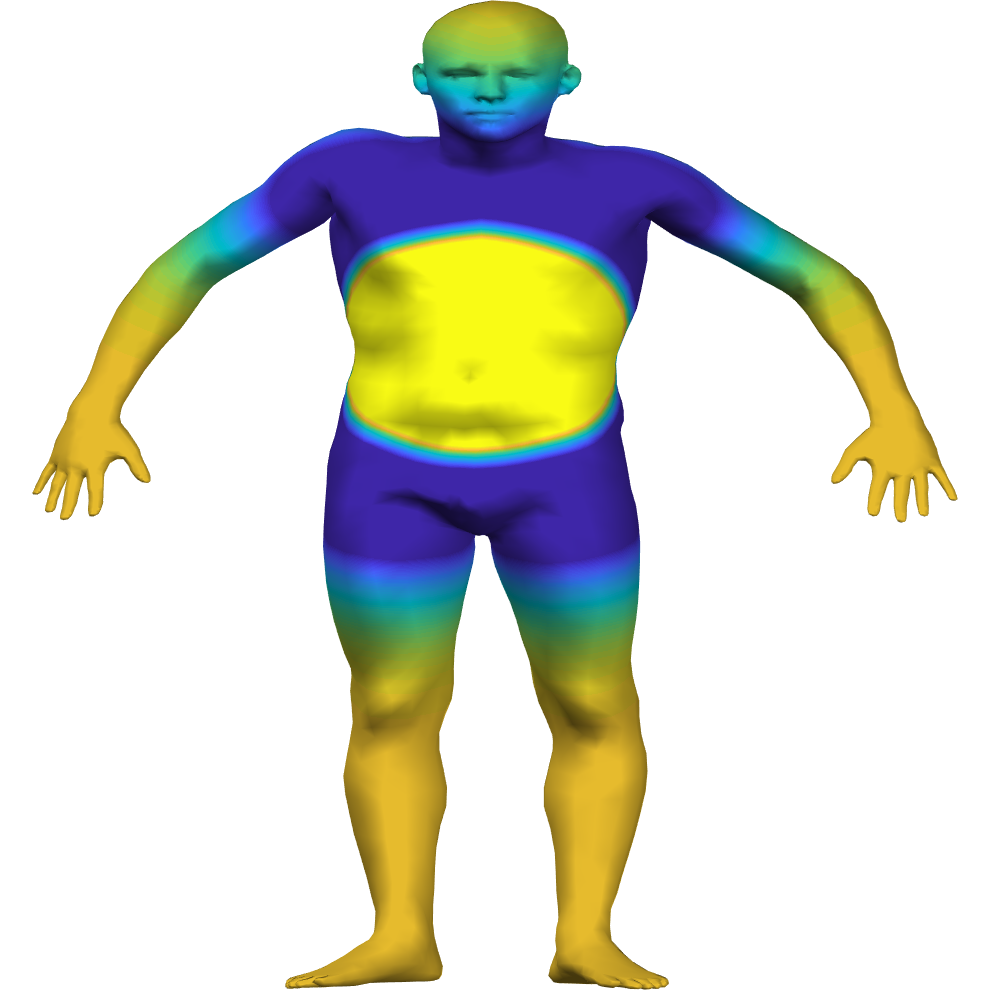}
\caption{}
\end{subfigure}
\begin{subfigure}{0.2\textwidth}
\includegraphics[scale=0.1]{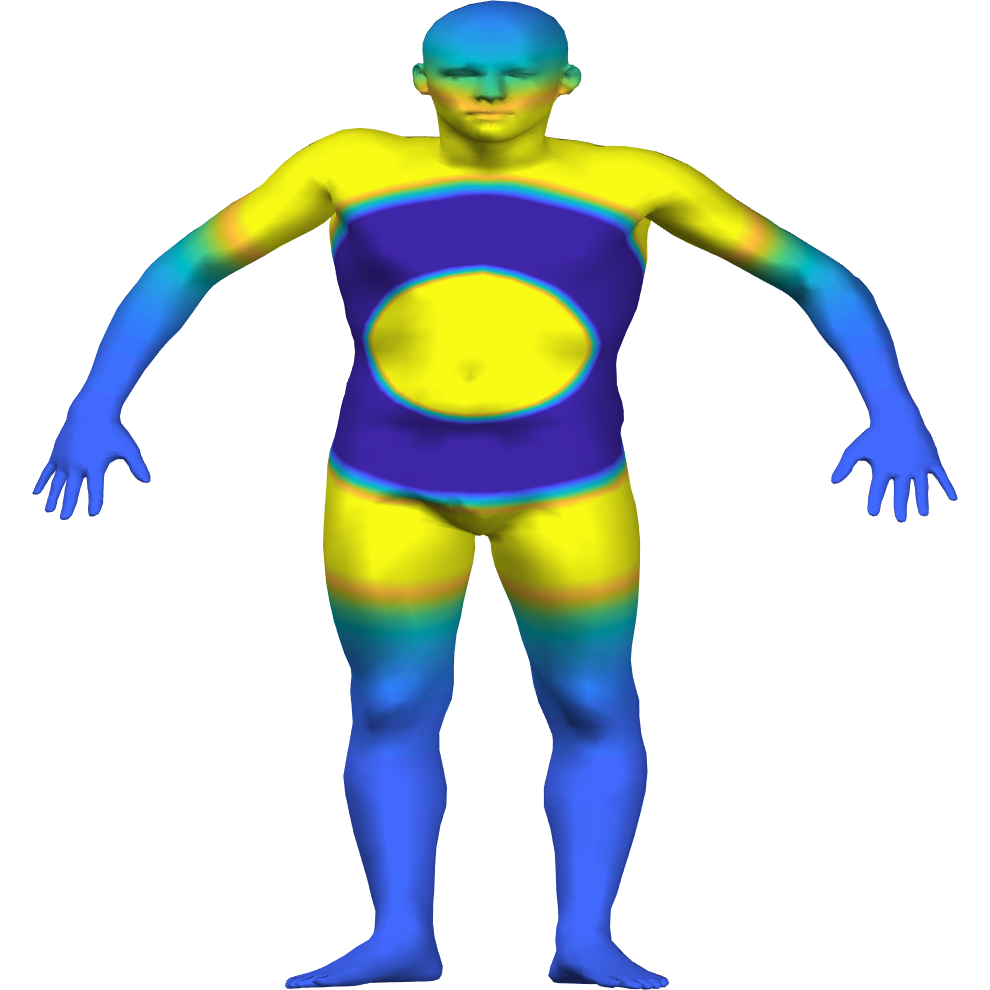}
\caption{}
\end{subfigure}
\begin{subfigure}{0.2\textwidth}
\includegraphics[scale=0.1]{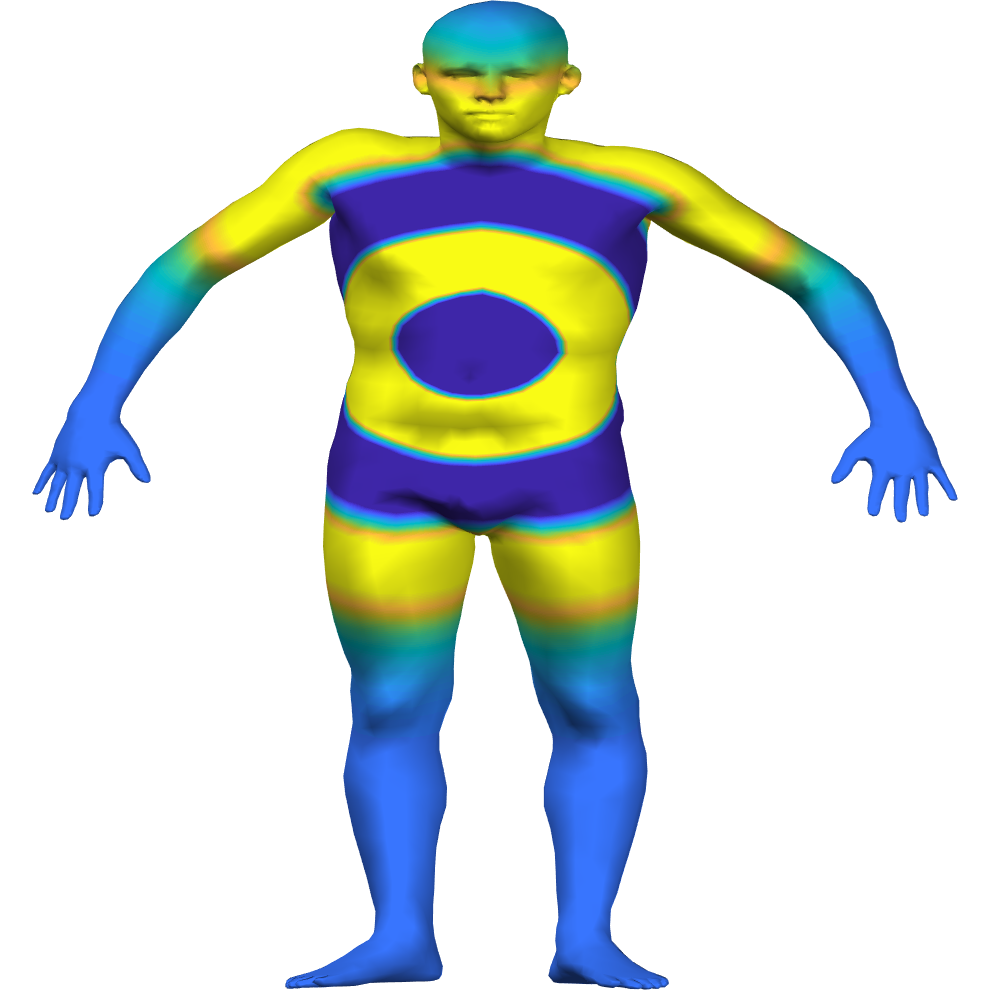}
\caption{}
\end{subfigure}
\end{center}
\caption{{\bf Eigenfunctions of the descriptor kernel of equation \ref{eq:kernel_equation_X}:}
(a) A gaussian kernel centered around a point on the belly. (b)-(e) second - fifth eigen-functions of the kernel operator. The colormap has been deliberately tresholded to convey the harmonic progression of the eigenvectors in the descriptor space. This shows a sequential clustering of regions with equal descriptor values. Enforcing the correspondence to commute with this kernel is equivalent to the additional constraint that these regions must also be preserved by the correspondence map, in addition to a straightforward point-wise preservation of the pointwise descriptor.}
\label{fig:kernel_eigfunc}
\end{figure*}
\section{Bilateral Operators}
Our main observation is that we can formulate a more informative pairwise constraint on the correspondence map from an input of point-wise descriptors. This means that in addition to requiring that the correspondence map preserve the descriptor values point-wise as in equation \ref{pointwise}, the correspondence map must also preserve the similarities between every pair of points as measured by the descriptor itself. For a given descriptor $f_{\mathcal{X}}^{(i)}$  and a pair of points $(x,x') \in \mathcal{X}$, we can associate a notion of similarity between them given by:
\begin{equation}
K_{f_{\mathcal{X}}^{(i)}}(x,x') = e^{-\frac{(f_{\mathcal{X}}^{(i)}(x)-f_{\mathcal{X}}^{(i)}(x'))^2}{2\sigma^2}}
\label{eq:kernel_equation_X}
\end{equation}
Equation \ref{eq:kernel_equation_X} is an operator constructed from a point-wise descriptor $f_{\mathcal{X}}^{(i)}$. One can also interpret it as a smoothing operator over the surface where the weights of averaging are determined by the kernel in equation \ref{eq:kernel_equation_X}. 

A more intrinsic notion of similarity over the surface is given by the heat kernel. Heat kernels are fundamental solutions to the heat diffusion equation over the manifold. 
\begin{equation}
    \frac{\partial u(t, x)}{\partial t}=\Delta_{\mathcal{X}} u(t, x)
\end{equation}
with the initial conditions $ u(0,x) = u_0(x)$. Here $u :[0, \infty) \times \mathcal{X} \rightarrow \mathbb{R}$ represents the amount of heat at point $x$ at time $t$. The solution to is given by:
\begin{equation}
    u(t, x)=\int_{\mathcal{X}} H_{\mathcal{X}}\left(t, x, x^{\prime}\right) u_{0}\left(x^{\prime}\right) d x^{\prime}
\end{equation}

The heat kernel can be represented in terms of the spectral decomposition of the LBO and is given by: 
\begin{equation}
    H_{\mathcal{X}}\left(t, x, x^{\prime}\right)=\sum_{i} e^{\lambda_{i} t} \phi_{i}(x) \phi_{i}\left(x^{\prime}\right)
    \label{eq:heat_kernel}
\end{equation}
For every pair of pointwise descriptors $\{f_{\mathcal{X}}^{(i)},f_{\mathcal{Y}}^{(i)}\} $, we can compute the corresponding bilateral operators that combine the descriptor similarity measure of \ref{eq:kernel_equation_X} with an intrinsic notion of similarity encoded by the heat kernel: 
\begin{eqnarray}
O_{\mathcal{X}}^{(i)}(x,x') &=& H_{\mathcal{X}}\left(t, x, x^{\prime}\right) + \gamma K_{f_{\mathcal{X}}^{(i)}}(x,x') \nonumber \\
O_{\mathcal{Y}}^{(i)}(y,y') &=& H_{\mathcal{Y}}\left(t, y, y^{\prime}\right) + \gamma K_{f_{\mathcal{Y}}^{(i)}}(y,y')
    \label{eq:bilateral_kernel}
\end{eqnarray}

The bilateral operators of equation \ref{eq:bilateral_kernel} are conceptually related to a promonient filtering technique used for the problem of image de-noising. Image de-noising is the problem of recovering an image signal from its noisy estimate. As in the case of any noise removal problem, the common intuition is to employ some form of smoothing to the image in which the value of the image function at each pixel is a weighted average of its neighbors. The classical signal processing view of smoothing employs a gaussian weighted averaging of the the image as in equation \ref{eq:gauss_smooth}.  
\begin{equation}
    \widehat{I}(\mathbf{p}_i) \propto \sum_{j \in \mathcal{N}_i} e^{\frac{-||\mathbf{p}_i-\mathbf{p}_j||^{2}}{2 \sigma^{2}}} I(\mathbf{p}_j)
    \label{eq:gauss_smooth}
\end{equation}
Here $\mathbf{p}_i \in \mathbb{R}^2$ encodes the 2D spatial location of pixel $i$, and $\mathcal{N}_i$ refers to the pixel indices in the spatial neighborhood of the $i^{th}$ pixel. $\widehat{I}$ is the estimate of the de-noised image. The operation of Equation \ref{eq:gauss_smooth} is equivalent to a heat diffusion of on the standard 2D euclidean plane with the noisy image $I$ as the initial condition. This is validated by the fact that the heat kernel in planar domain is the standard gaussian kernel:
\begin{equation}
    H_{2D}(t,\mathbf{p}_i, \mathbf{p}_j)=\frac{1}{4 \pi t} e^{-||\mathbf{p}_i-\mathbf{p}_j||^{2} / 4 t}
    \label{eq:heat_2D}
\end{equation}

However, an important property desirable from such smoothing operators is edge preservation. See figure \ref{bilateral_image}. A gaussian smoothing of equation \ref{eq:gauss_smooth} will smooth away essential details of the image like dominant edges and textures, since the weights are independent of image intensity. 
This motivated certain edge preserving filters that allowed for a image-dependent heat diffusion. The beltrami flow framework \cite{sochen2001diffusions,spira2007short} and the the bilateral filter \cite{tomasi1998bilateral} introduced an additional component that weights a pair of pixels by the difference in their image values in addition to geometric closeness ensured by the heat kernel. Therefore the action of the bilateral filter was to average pixels that were close in the geometric and functional domain, leading to an improved image de-noising filter.     
\begin{equation}
    \widehat{I}(\mathbf{p}_i) \propto \sum_{j \in \mathcal{N}_i} \{ e^{\frac{-||\mathbf{p}_i-\mathbf{p}_j||^{2}}{2 \sigma^{2}}} \; . \;   e^{\frac{-|I(\mathbf{p}_i)-I(\mathbf{p}_j)|^{2}}{2 \beta^{2}}}  \} I(\mathbf{p}_j)
    \label{eq:bilateral_smooth}
\end{equation}

In this paper, we assert that a similar argument can be extended to the recovery of correspondence maps. Given the knowledge of good quality point-wise descriptors, we assert that a desirable correspondence map must preserve a combination of both geodesic and descriptor closeness. This can be achieved by commuting with the bilateral operators of equation \ref{eq:bilateral_kernel}. 

There is however a minor conceptual difference between the bilateral operators used in image smoothing and the operators of equation \ref{eq:bilateral_kernel}. An edge preserving kernel for image smoothing demands that the points be spatially close and also have similar image intensities. Hence the resulting bilateral operator is a pointwise multiplication of the kernels (or a Hadamard multiplication of two positive definite matrices). However, this property is appropriate for a de-noising scenario where both similarities need to be strong for the final weight to be significant. 

In contrast, the bilateral operator proposed in this paper suggests an additive combination of geodesic and descriptor similarity. Hence either of the similarities need to be strong for the final kernel value to be significant. 
The constraint of commuting with such an operator leads to the following observation. Pairs of points on the source shape that are similar in a geodesic sense (hence a higher value for the heat kernel) or by way of having identical descriptor values must correspond distinctly to similar pairs on the target. The optimal correspondence ignores only those pairs that are neither geodesically close nor have similar descriptor values. Intuitively, this policy works favourably when the shapes are not strictly related by an isometry, since there exist some pairs of points that are relatively far apart on the surface but worthy of good matching. The heat kernel may not weight such pairs highly. However, such pairs can be rightly deemed close by an informative pointwise descriptor and this relationship is leveraged by the pairwise bilateral operator. Hence commuting with the bilateral operator enables a more informative utilization of each descriptor leading to a better correspondence quality for a smaller budget of point-wise descriptors.

Figure \ref{fig:kernel_eigfunc} provides a visualization on the action of the kernel operator of equation \ref{eq:kernel_equation_X}. The eigenfunctions show a harmonic progression in the descriptor space, that is, the shape is harmonically clustered into regions of similar descriptor values. This is in contrast to the global manifold harmonics that are the eigenfunctions of the heat kernel. Therefore, given that we have knowledge of a pair of corresponding descriptors, we require that, not only must the correspondence map preserve the value of the descriptor pointwise, but also the layered pairwise relationship between every pair of points as encoded by the descriptor centric kernel of Equation \ref{eq:kernel_equation_X}. This philosophy works best when we combine it with a geometric notion of similarity encoded in the heat kernel. 
\subsection{Numerical Computation}
\label{sec:bilateral_numericalcomp}
For each descriptor, an explicit construction of the bilateral operators $\{O_{\mathcal{X}}^{(i)},O_{\mathcal{Y}}^{(i)} \in \mathbb{R}^{n \times n}\}$ demands the population of a $n \times n $ matrix using equation \ref{eq:kernel_equation_X}. However, for a functional map estimation, we only need the spectral equivalents of the operators $\{\widehat{O}_{\mathcal{X}}^{(i)},\widehat{O}_{\mathcal{Y}}^{(i)} \in \mathbb{R}^{k \times k} \}$ for its use in the optimization \ref{eq:fmap_energymin}. These can be computed as follows:
\begin{eqnarray}
\widehat{O}_{\mathcal{X}}^{(i)} &=& \begin{bmatrix} e^{\lambda_1 t} & &\\& \ddots & \\& & e^{\lambda_k t}\end{bmatrix} + \Phi^T \mathbf{K}_{f_{\mathcal{X}}^{(i)}}\Phi \nonumber \\
\widehat{O}_{\mathcal{Y}}^{(i)} &=& \begin{bmatrix} e^{\mu_1 t} & &\\& \ddots & \\& & e^{\mu_k t}\end{bmatrix} + \Psi^T \mathbf{K}_{f_{\mathcal{Y}}^{(i)}}\Psi
\label{eq:bilateral_spectral}
\end{eqnarray}
For spectral projection of the descriptor kernels $\Phi^T \mathbf{K}_{f_{\mathcal{X}}^{(i)}}\Phi$ and $\Psi^T \mathbf{K}_{f_{\mathcal{Y}}^{(i)}}\Psi$, we use the nystrom extension for approximating large positive definite matrices \cite{williams2001using}. In particular, the Nystrom approximation requires only a sparse set of columns of any kernel matrix. Let $\mathbf{R} \in \mathbb{R}^{n \times n_0},\; n_0 \ll n$ be such a column stacked matrix of the columns of $\mathbf{K}_{f_{\mathcal{X}}^{(i)}}$. Let $\mathbf{R_0} \in \mathbb{R}^{n_0 \times n_0}$ be the smaller intersecting submatrix of $\mathbf{R}$. Then we have the following expressions: 
\begin{eqnarray}
    \mathbf{K}_{f_{\mathcal{X}}^{(i)}} &\approx& \mathbf{R} \; \mathbf{R_0}^{-1} \; \mathbf{R}^T \\
    \Phi^T \mathbf{K}_{f_{\mathcal{X}}^{(i)}} \Phi &\approx& (\Phi^T \mathbf{R}) \; \mathbf{R_0}^{-1} \; (\Phi^T\mathbf{R})^T 
    \label{eq:nystrom}
\end{eqnarray}
and a similar expression may be obtained for the kernel matrices for shape $\mathcal{Y}$. We use the farthest point sampling strategy \cite{hochbaum1985best} for constructing the column matrix $\mathbf{R}$. By evaluating the kernel functions \ref{eq:kernel_equation_X} for only a sparse subsampled set of points and use the approximations of equation \ref{eq:nystrom} we arrive at the spectral coefficients for the pairwise bilateral operators to be used in a functional map pipeline.  

\section{Correspondence Algorithm}
\begin{algorithm}[H]
\begin{tabular}{ll}
{\bf Input:}& Source and Target shapes $\mathcal{X},\mathcal{Y}$,\\ 
&spectral geometries of shapes : $\Delta_{\mathcal{X}},\Delta_{\mathcal{Y}}$ \\ 
&$\Phi, \Psi, \Lambda_{\mathcal{X}}, \Lambda_{\mathcal{Y}}$ \\
{\bf Output:}& Functional Map matrix C,\\
&correspondence map $\varphi$\\
\end{tabular}\\
\begin{enumerate}
\setlength\itemsep{0.7em}
\item Find pairs of pointwise descriptors for each shape and compute their spectral equivalents $\{ \widehat{f}_{\mathcal{X}}^{(i)}, \widehat{f}_{\mathcal{Y}}^{(i)}\}_{i}$ 
\item From each pair, construct the corresponding bilateral operators   $\{\widehat{O}_{\mathcal{X}}^{(i)}, \widehat{O}_{\mathcal{Y}}^{(i)}\}_{i}$ using the computational routine outlined in section \ref{sec:bilateral_numericalcomp}
\item Incorporate constraints into a linear system and solve for functional map matrix C by minimizing equation \ref{eq:fmap_energymin} 
\item Refine the map using an ICP like iteration \cite{ovsjanikov2012functional}
\item Compute a point to point correspondence using the refined map
\end{enumerate}
\caption{Bilateral Operators For Functional Maps}
\label{algo_bilateral_fmap}
\end{algorithm}
\section{Experiments}
The general philosophy of our experiments is to compare between various functional map algorithms by imputing all of them with the \emph{same} set of descriptors and using the same linear optimizer with identical coefficients for the pairwise and pointwise constraints. 
We compare with three algorithms: the original functional maps framework \cite{ovsjanikov2012functional} without any descriptor based regularizer, the diagonal operator proposed in \cite{nogneng2017informative} and kernel functional maps \cite{wang2018kernel}. 
All of these algorithms have the same objective of maximizing the information from a fixed budget of pointwise descriptors.

First, we conduct a synthetic experiment to assess the variation in correspondence quality as a function of number of \emph{perfect} input pointwise descriptors over 25 shape pairs of FAUST \cite{bogo2014faust}. The objective of this experiment is to assess which functional map algorithm (or constraint) maximally exploits the information from a fixed budget of high quality descriptors. Most recent and arguably state-of-the-art algorithms on shape correspondence are tailored towards learning optimal descriptors with properties of localization, sensitivity and invariance. Recent results from deep-learning frameworks have demonstrated that such descriptors can be learned from examples \cite{litman2013learning,boscaini2016learning,litany2017deep,Halimi_2019_CVPR}. What we intend to answer with the first experiment is the question, \emph{Given that we have high-quality pointwise descriptors, which functional map algorithm achieves the best correspondence quality?}
\begin{figure*}
\centering
\includegraphics[scale=0.4]{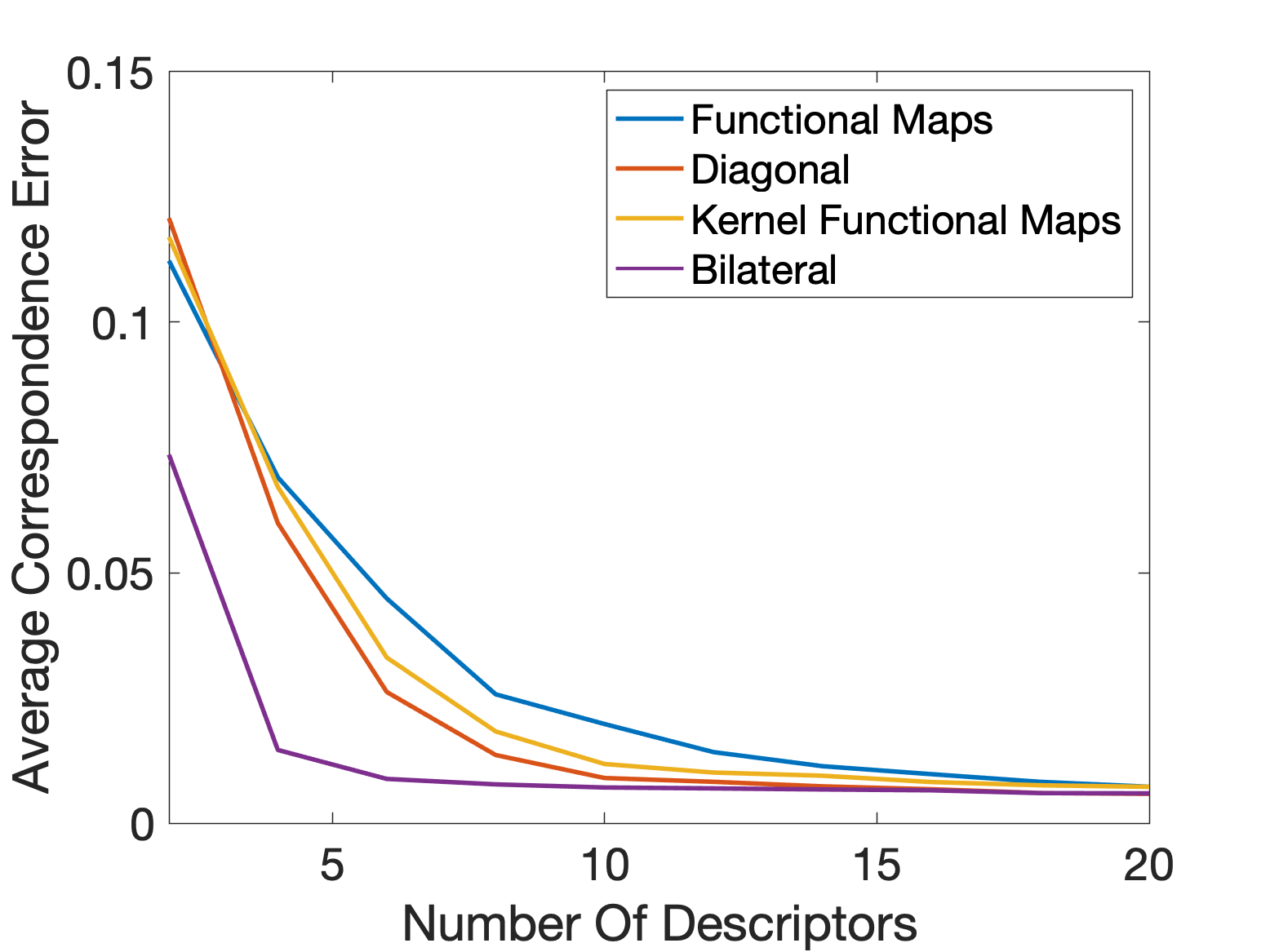}
\caption{Correspondence quality vs number of descriptors for various algorithms.}
\label{desc_errorplot}
\end{figure*}
We compute descriptors by constructing gaussian kernels over ground-truth corresponding points (see Figure \ref{fig:kernel_eigfunc} (a)), and progressively increase the number of such descriptors as the input to the different algorithms. 
\begin{equation}
    f_{\mathcal{X}}^{(i)} = e^{\frac{-d(x_i,x)^2}{\sigma^2}}
\end{equation}
What we expect to see is a decay in the average correspondence geodesic error as the number of such descriptors increase for each algorithm. We use 60 eigenbasis functions for all shapes algorithms. The parameters for all algorithms we chosen as per the choices made in the respective papers. For the bilateral kernel we chose $t=1e-3$ and $\sigma = 3$ and $\gamma = 1$ for all experiments.  Figure \ref{desc_errorplot} shows the results.  We see that for a few descriptors, the bilateral filter shows a much lower average correspondence error for the same number of input descriptors. However, all algorithms eventually converge similarly when the number of descriptors increase. Table \ref{fig:desc_map} visualizes the correspondence for the first few landmarks of the plot. 

However, general descriptors are far from perfect. This means that there will be a problem of noise leading to inaccurate correspondence. Therefore, in such scenarios, it is not surprising to expect that increasing the number of descriptors may not necessarily lead to an improved correspondence. Moreover the existence of outliers may even deteriorate correspondence quality.  

\begin{figure*}[t]
\centering
\includegraphics[width=1\textwidth]{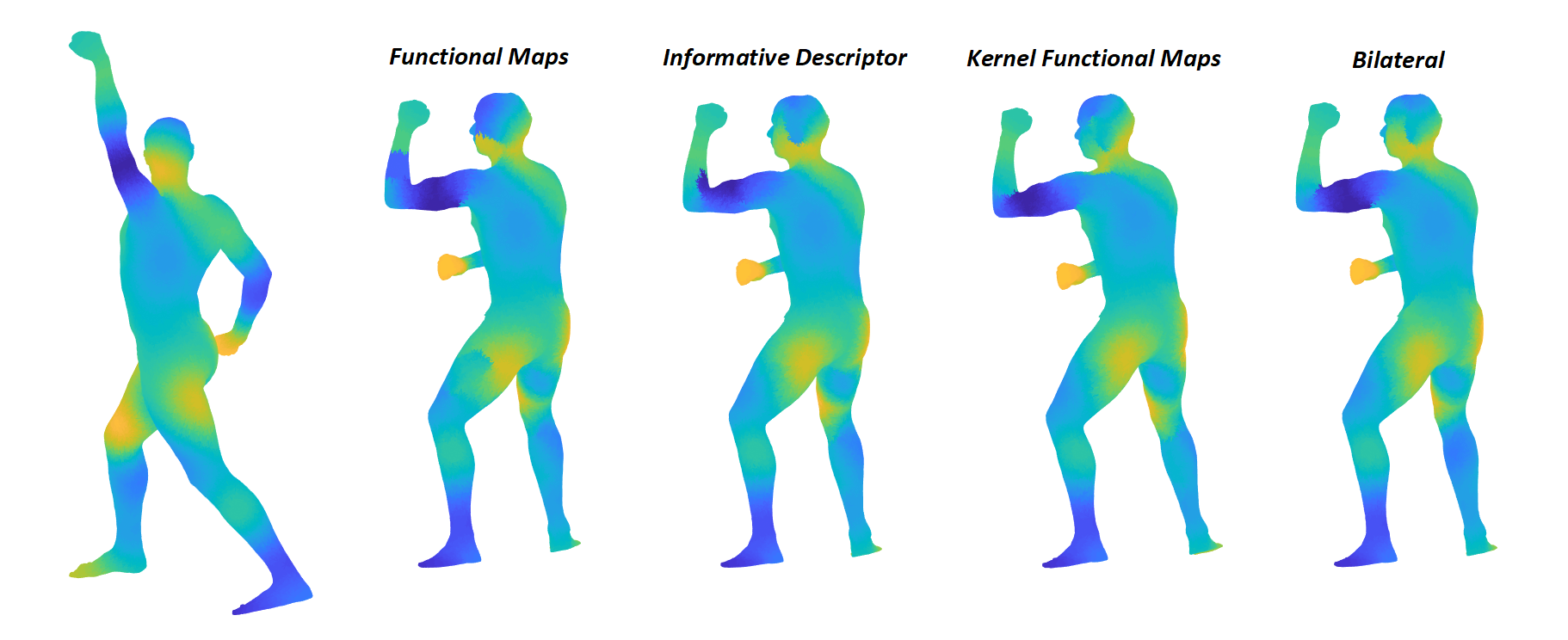}
\caption{Example of point to point mapping of smooth function on the source (left) to the target, obtained by different functional maps algorithms using 10 wave kernel descriptors. 
}
\label{p2p_map}
\end{figure*}

\begin{figure*}
\centering
\begin{center}
\hspace{-0.5cm}
\begin{subfigure}{0.49\textwidth}
\includegraphics[scale=0.4]{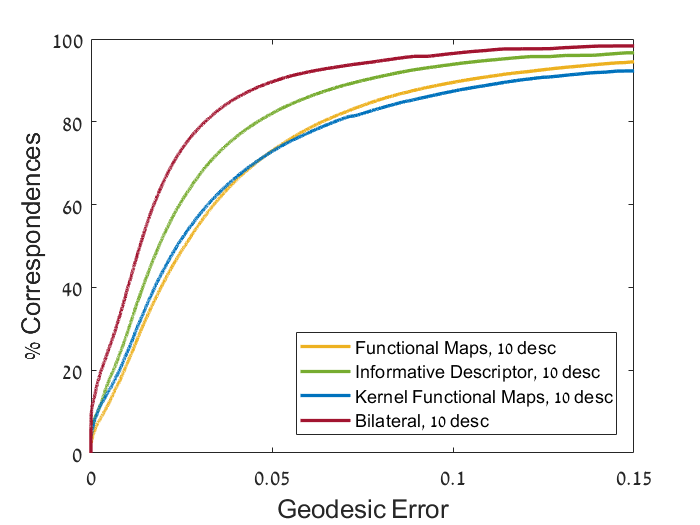}
\caption{Faust, 10 descriptors}
\end{subfigure}
\begin{subfigure}{0.49\textwidth}
\includegraphics[scale=0.4]{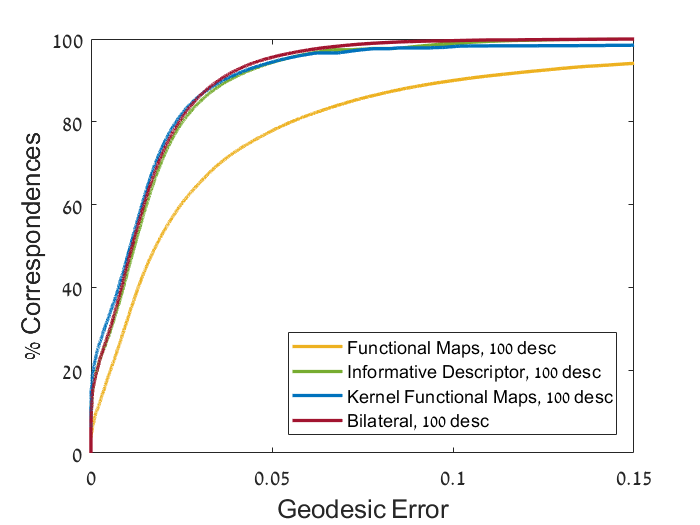}
\caption{Faust, 100 descriptors}
\end{subfigure}

\begin{subfigure}{0.49\textwidth}
\includegraphics[scale=0.4]{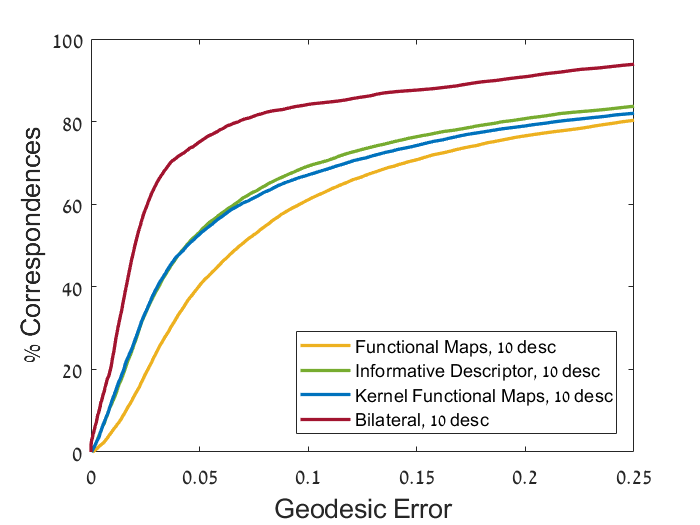}
\caption{Scape, 10 descriptors}
\end{subfigure}
\begin{subfigure}{0.49\textwidth}
\includegraphics[scale=0.4]{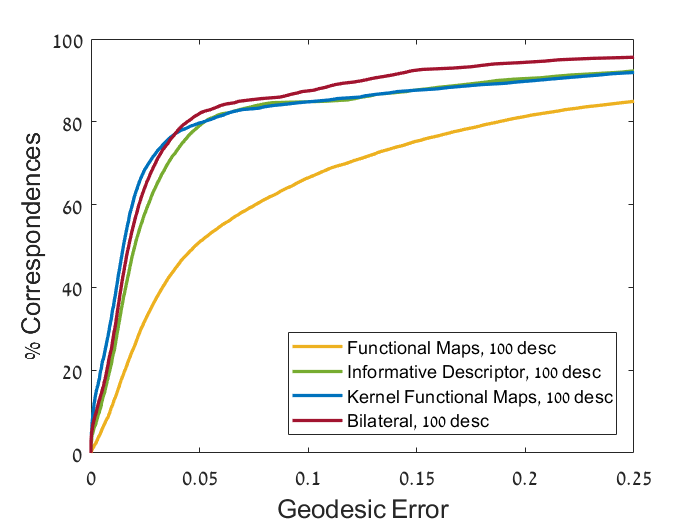}
\caption{Scape, 100 descriptors}
\end{subfigure}

\end{center}
\caption{Comparison of our method with Functional Maps \cite{ovsjanikov2012functional}, Informative Descriptor \cite{nogneng2017informative} and Kernel Functional Maps \cite{wang2018kernel} on two datasets: SCAPE \cite{anguelov2005scape} and FAUST \cite{bogo2014faust}.
FAUST \cite{bogo2014faust}. }
\label{quantitiave_results}
\end{figure*}

Figure \ref{quantitiave_results} shows the results of our approach compared to \cite{ovsjanikov2012functional},\cite{nogneng2017informative},\cite{wang2018kernel} on SCAPE and FAUST datasets. We evaluate each algorithm on randomly 100 chosen pairs, for each shape we used 100 discrete laplace-beltarami eigenfunctions as the functional basis. The results of figures \ref{quantitiave_results} show that the bilateral kernel is very potent at lower number of descriptors, and compares favourably with other algorithms for larger number of descriptors, similar to the results in of figure \ref{desc_errorplot}.  
Figure \ref{p2p_map} shows comparison of qualitative results of point to point mapping from the source to a target shape using 10 wave kernel signature/map \cite{aubry2011wave} descriptors. The qualitative results are inline with the quantitative ones, see figure \ref{quantitiave_results}.
\section{Conclusion}
In this paper, we propose a novel operator constructed from a pointwise descriptor. We introduce the bilateral operator that combines a descriptor based similarity with the intrinsic geometry of the surface encoded in the heat kernel. By enforcing commutativity constraints with this operator, we demonstrate that we can achieve a better correspondence quality, given a fixed budget of pointwise descriptors. 
\clearpage
\begin{figure*}[t]
\centering
\includegraphics[scale=0.23]{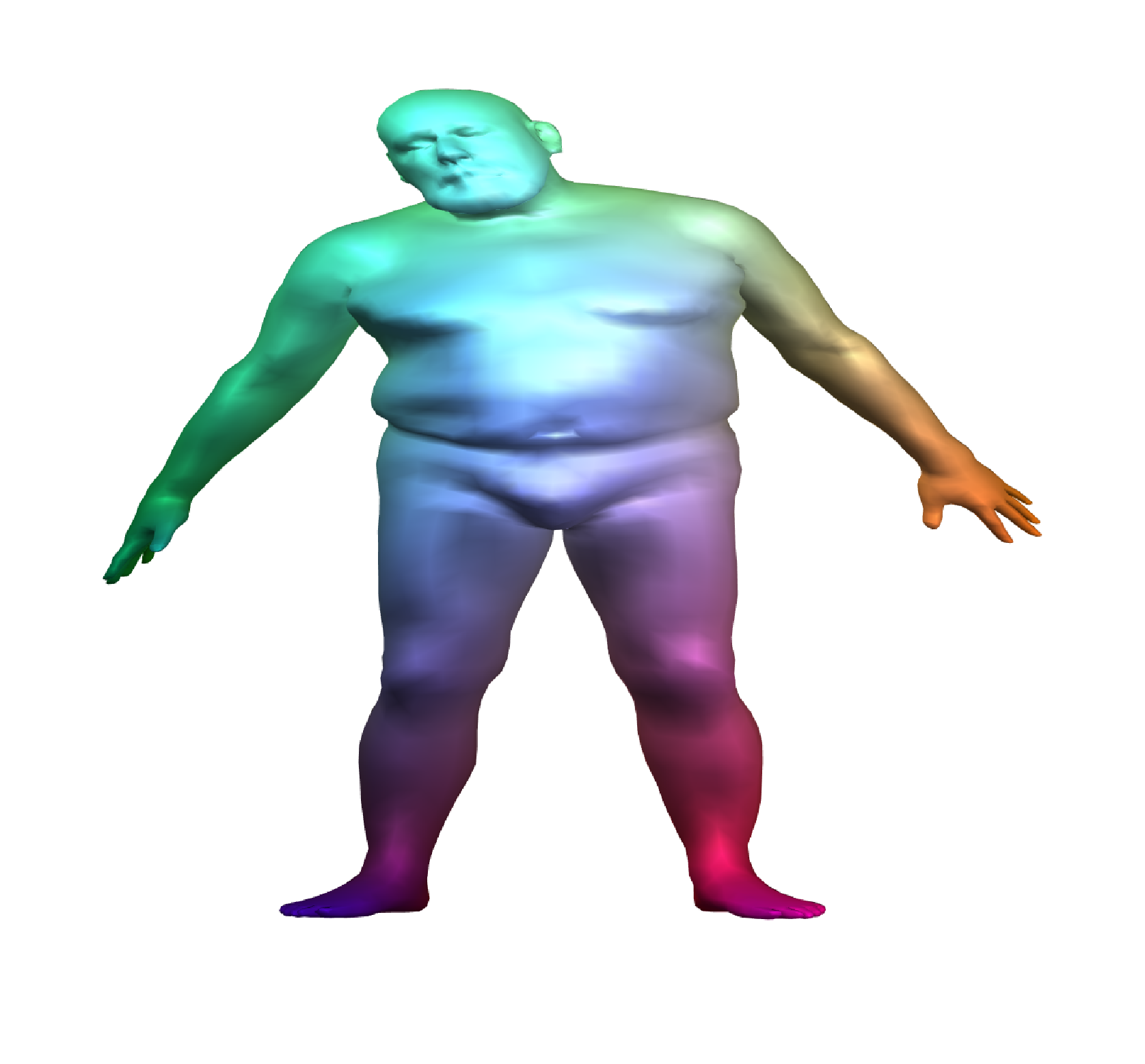}
\caption{A source shape}
\label{src_visual}
\end{figure*} 
\begin{table*}
\begin{tabular}{r*3{C}@{}}
 & 2 Descriptors & 4 Descriptors & 6 Descriptors \\ 
 Functional Map & \includegraphics[scale=0.09]{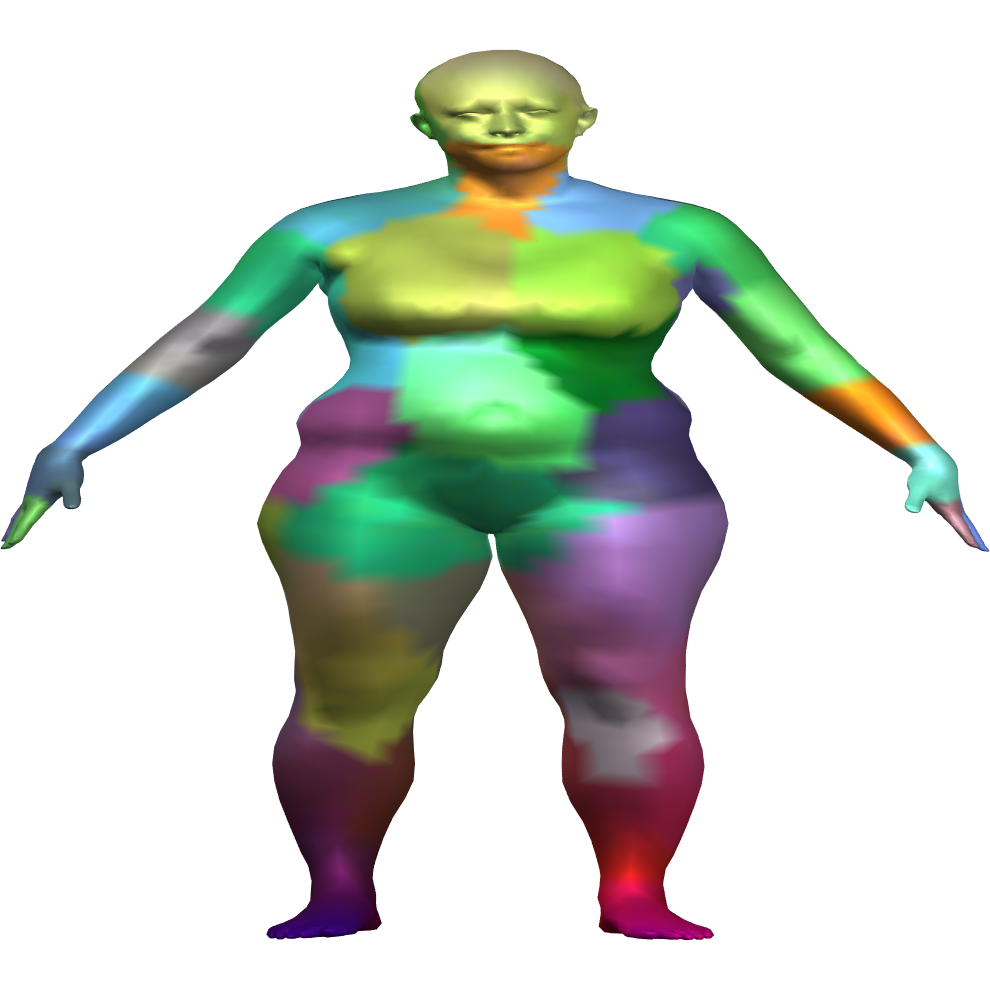} &  \includegraphics[scale=0.09]{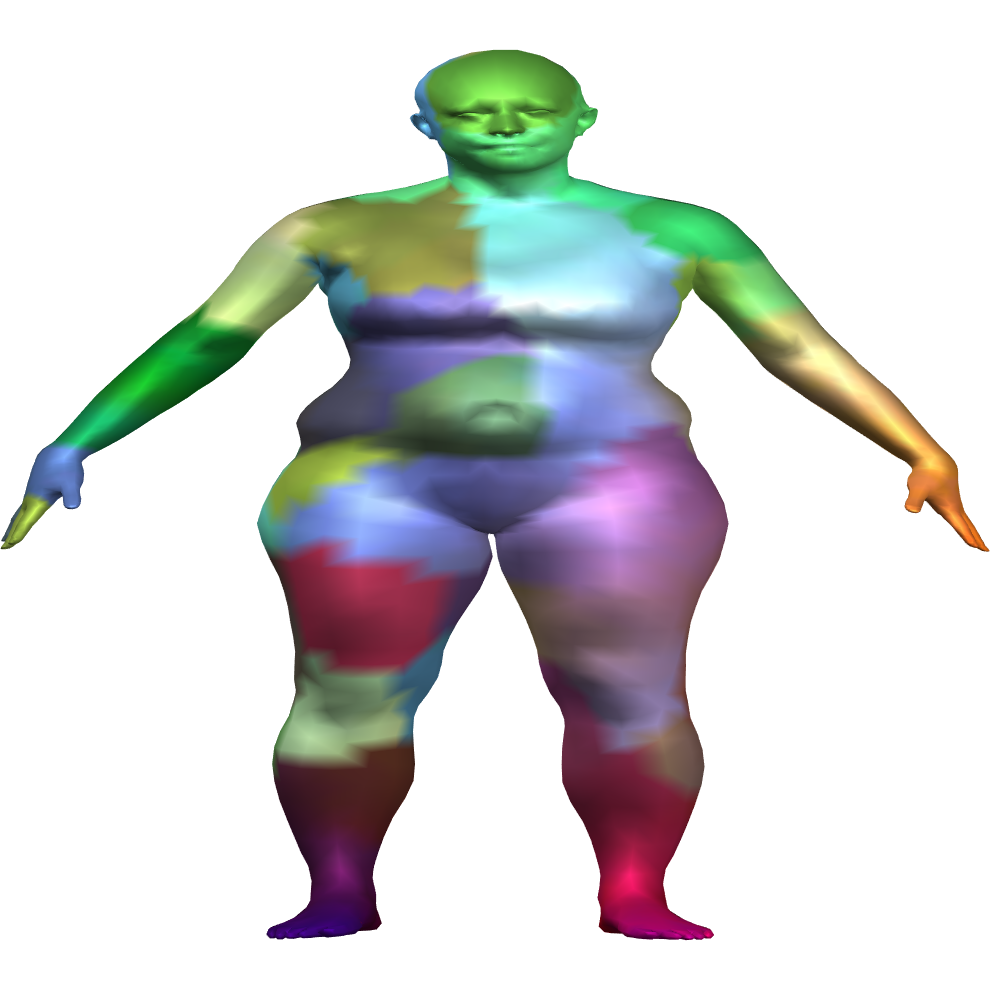} & \includegraphics[scale=0.09]{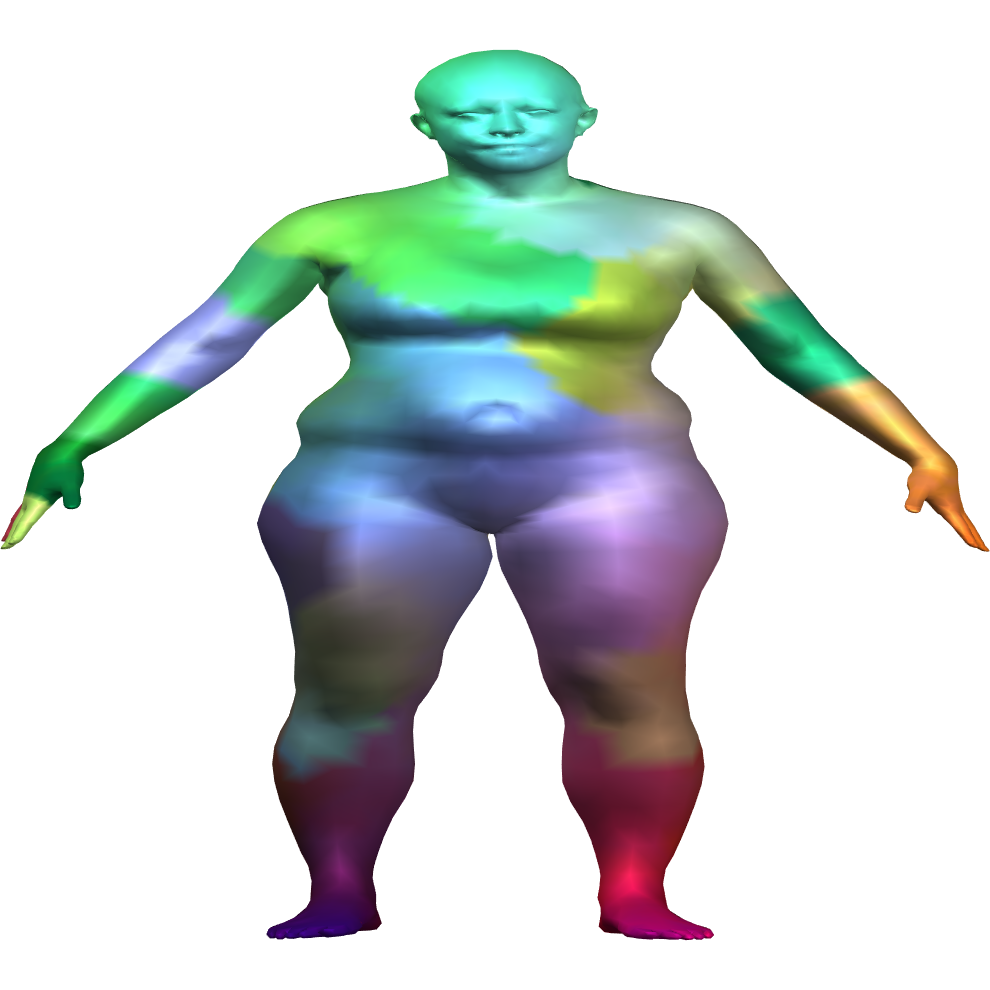} \\ 
 Diagonal & \includegraphics[scale=0.09]{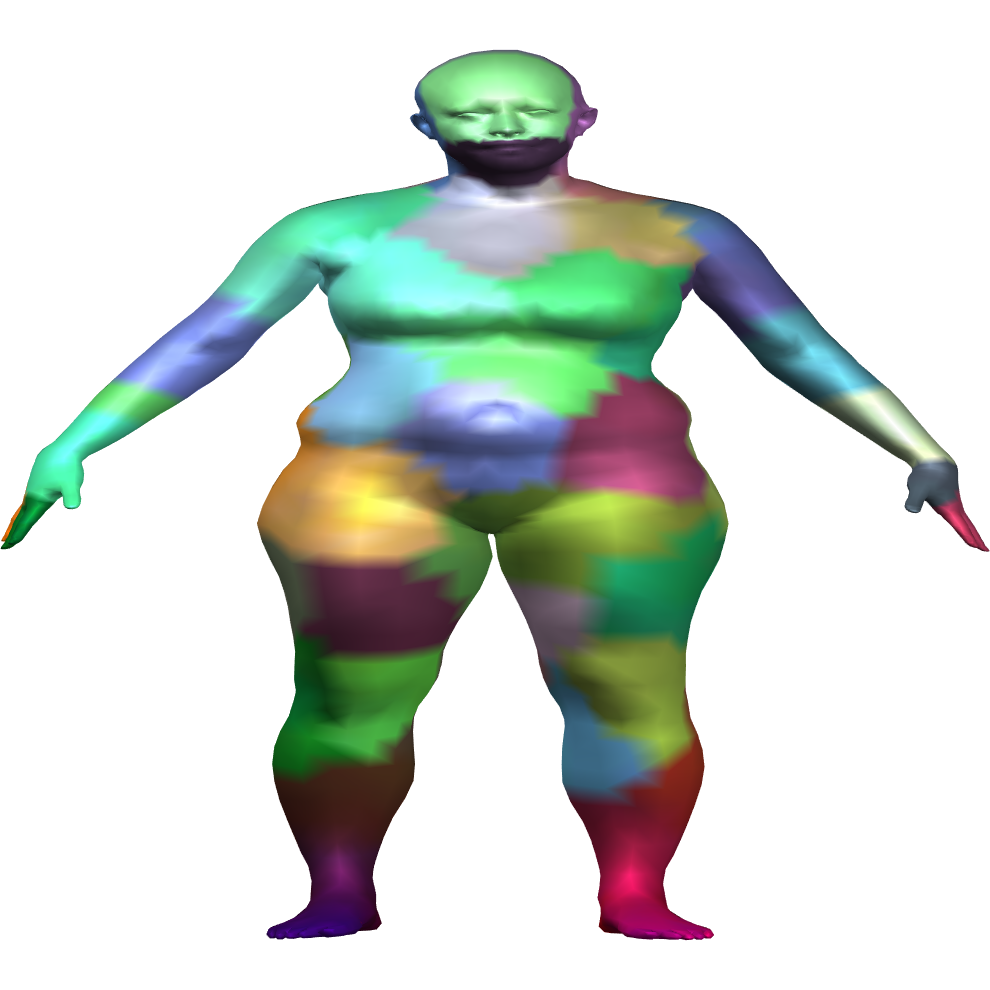} & \includegraphics[scale=0.09]{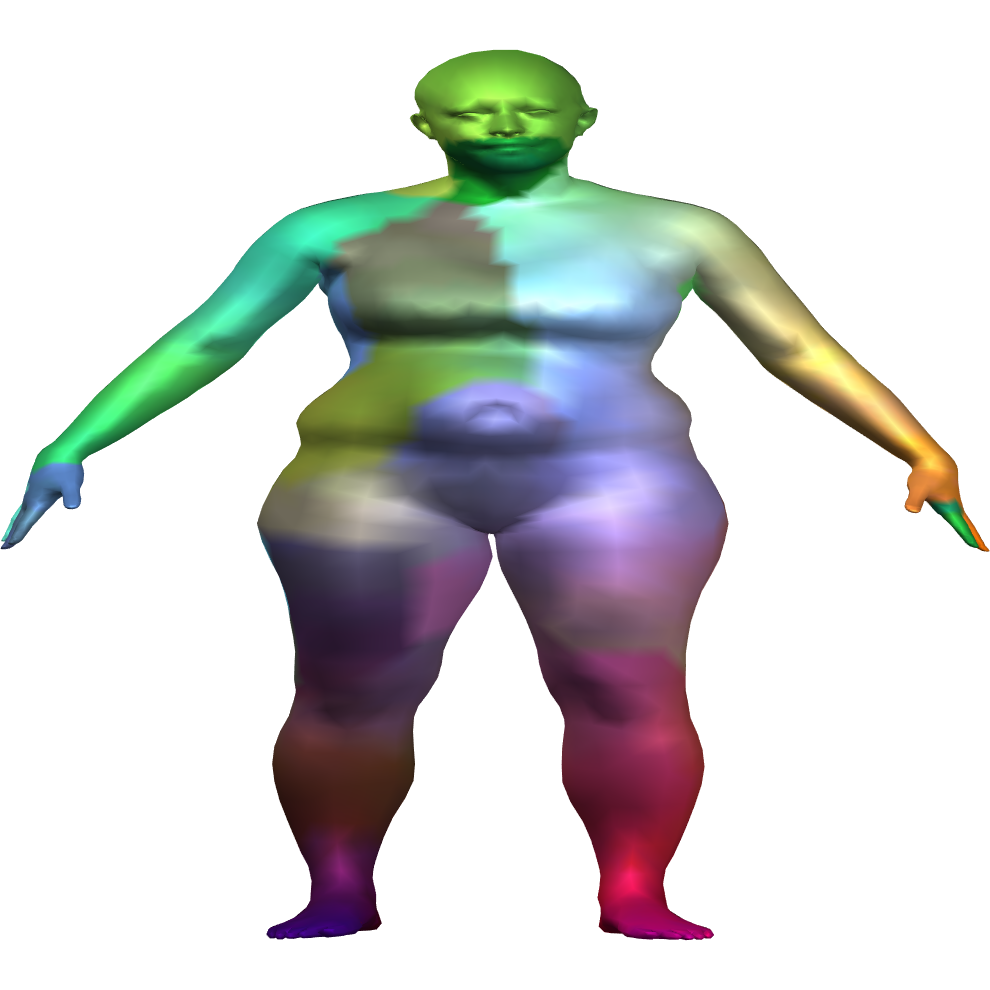} & \includegraphics[scale=0.09]{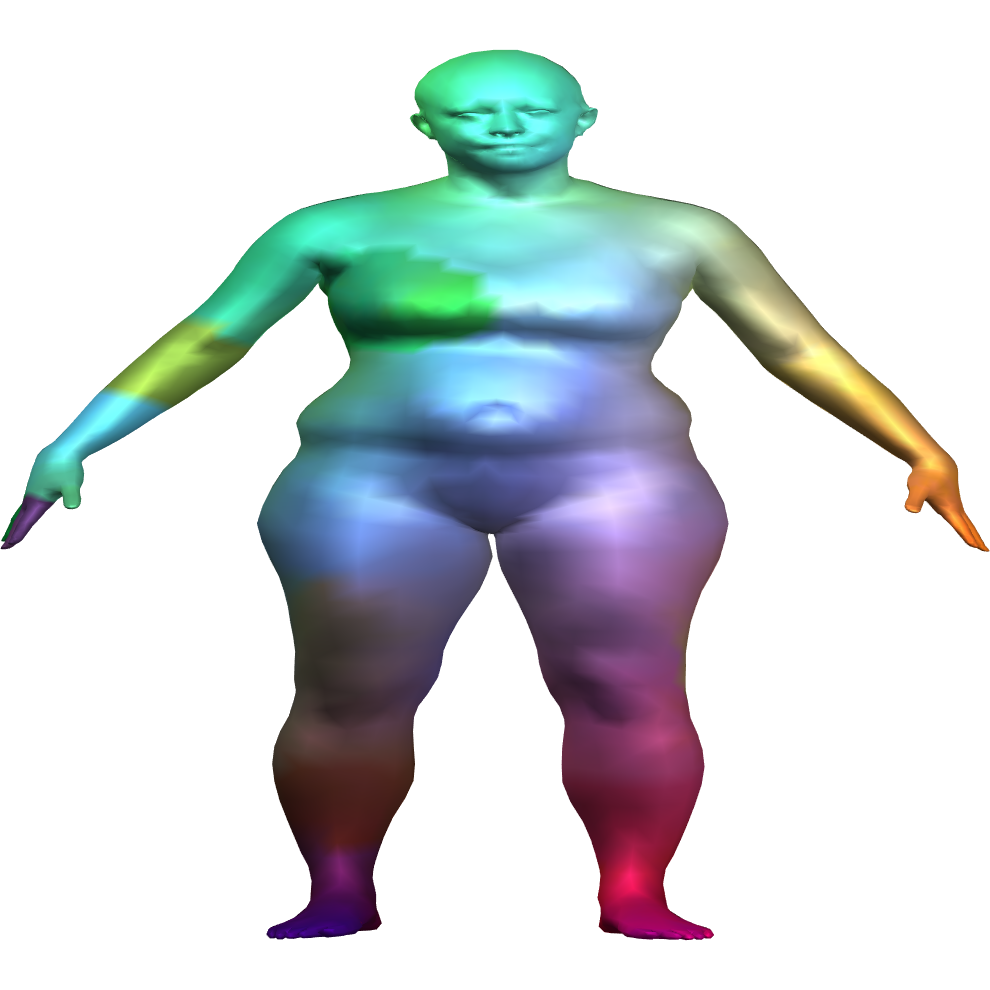} \\ 
  Kernel Functional Maps & \includegraphics[scale=0.09]{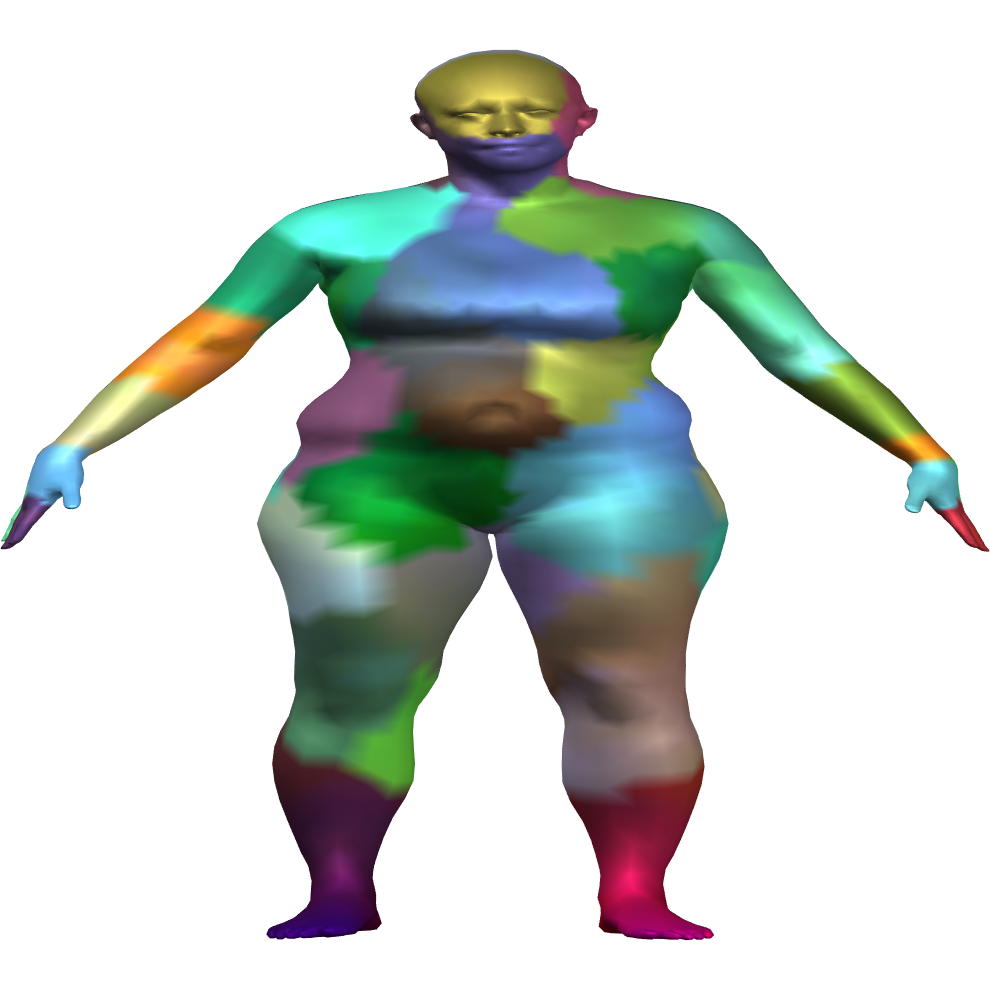} & \includegraphics[scale=0.09]{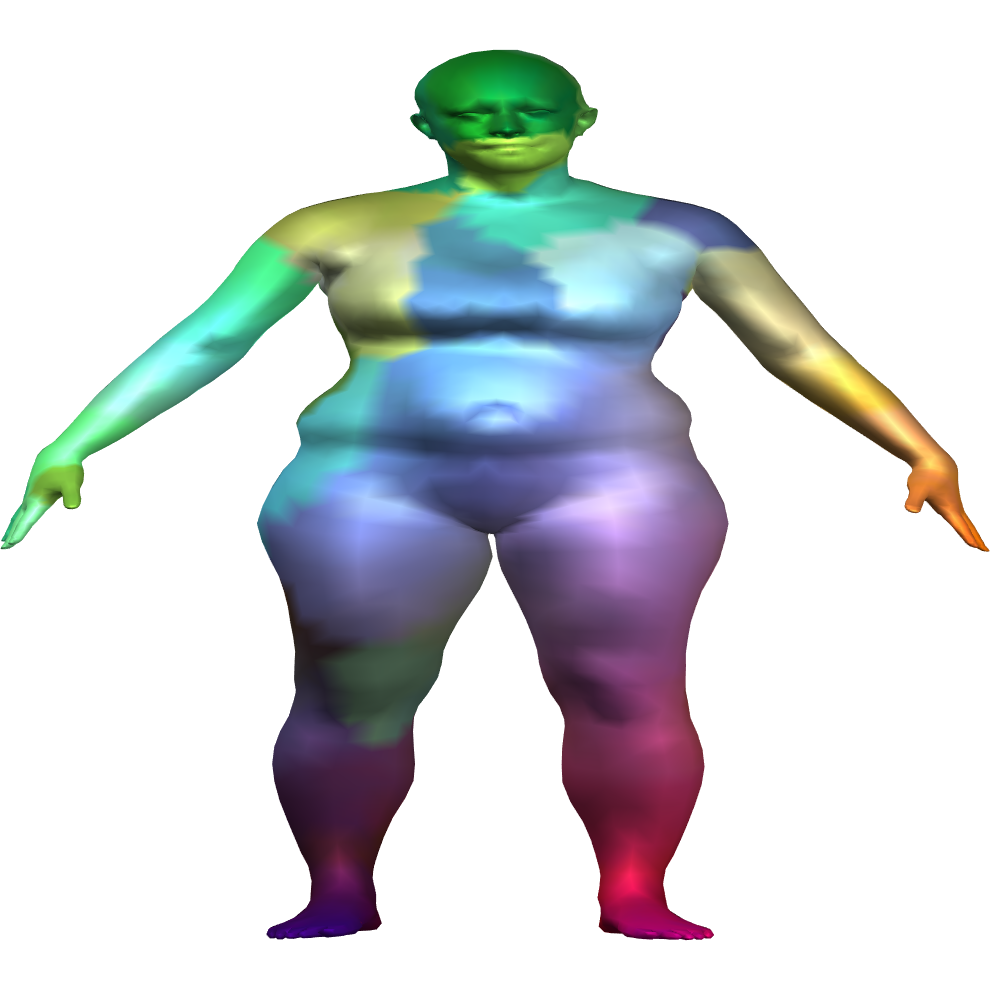} & \includegraphics[scale=0.09]{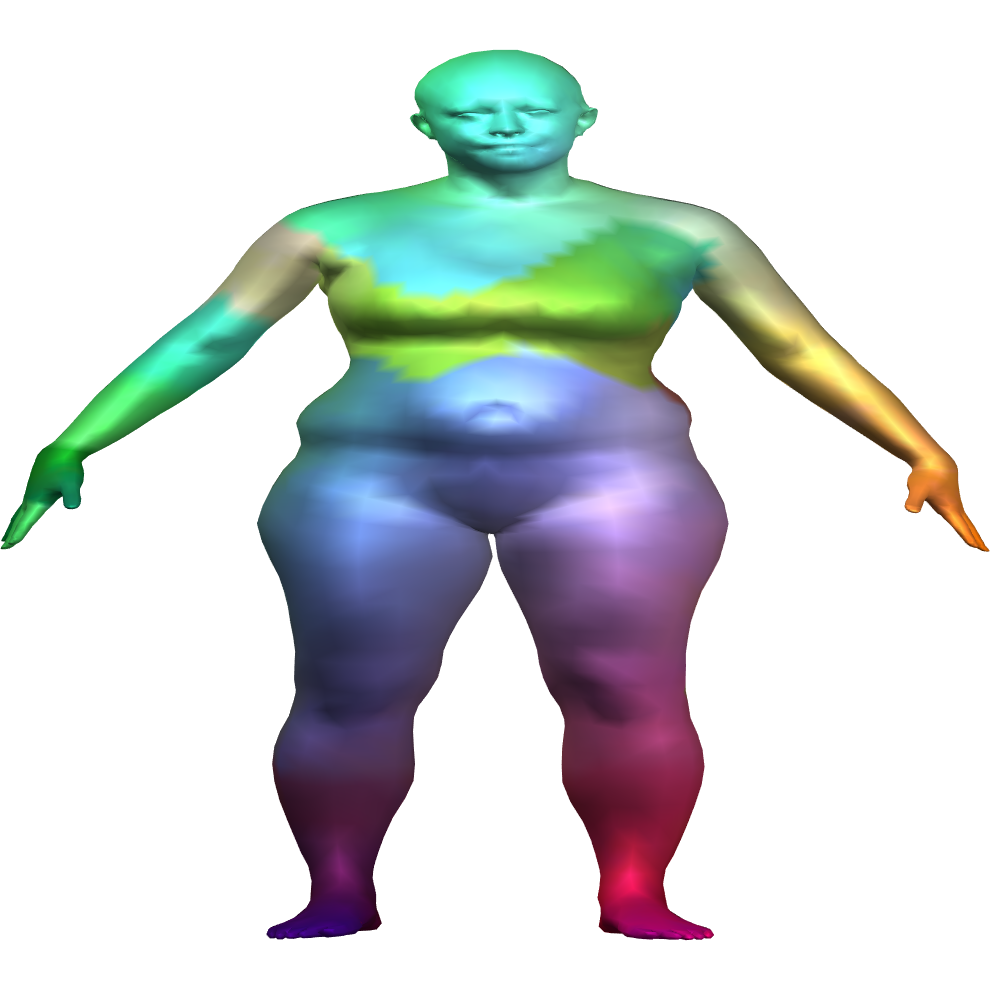} \\ 
 Bilateral & \includegraphics[scale=0.09]{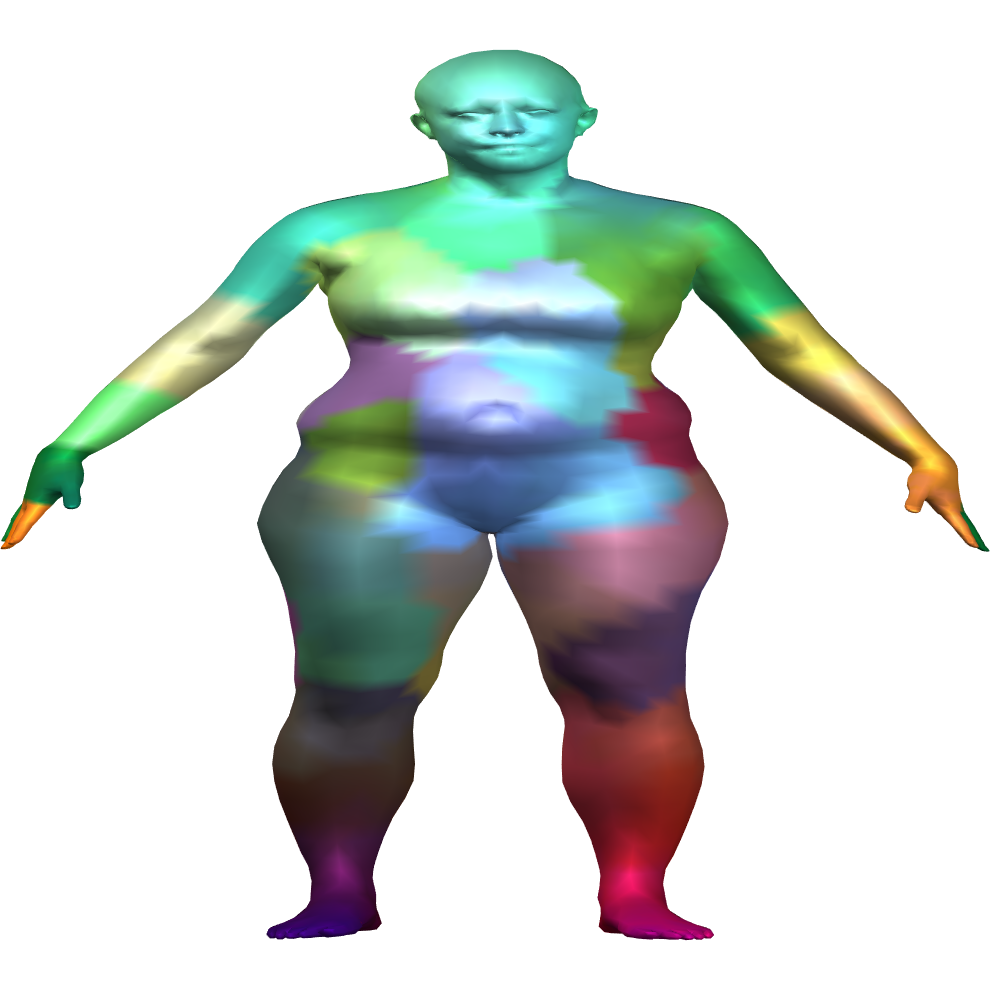} & \includegraphics[scale=0.09]{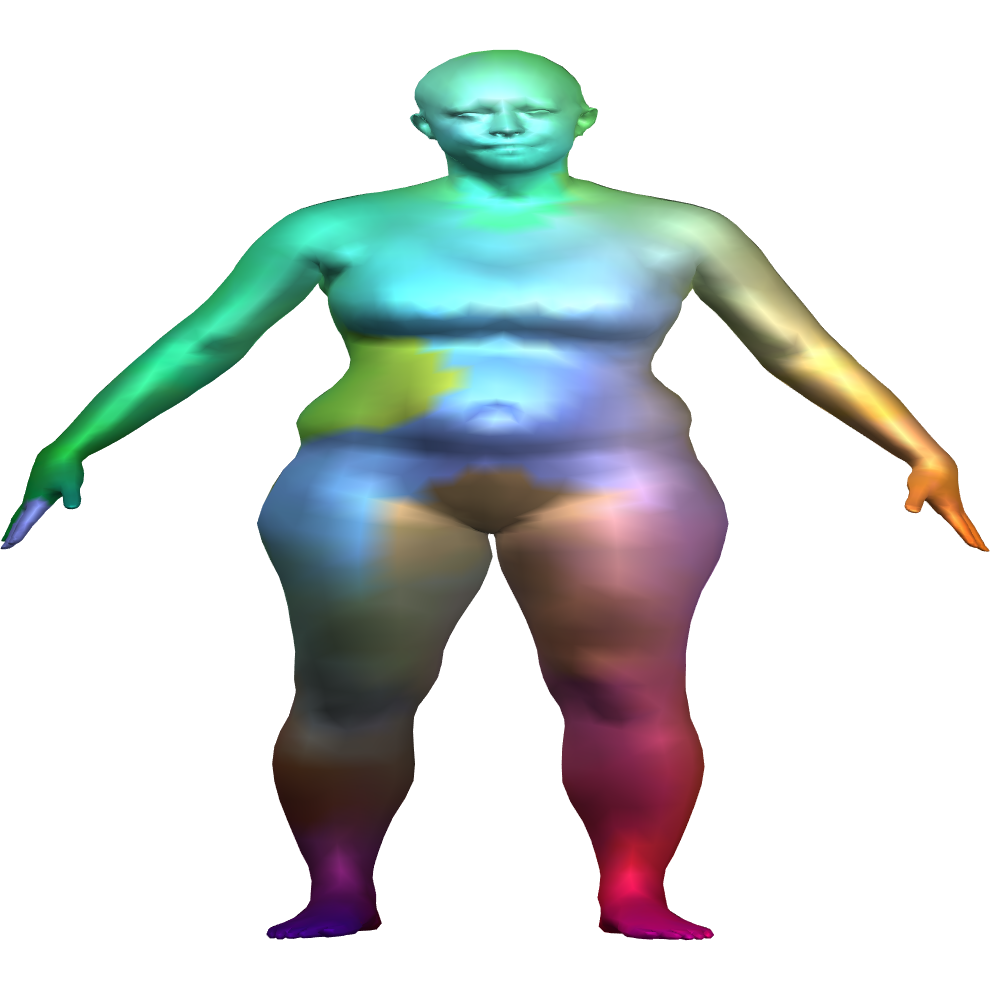} & \includegraphics[scale=0.09]{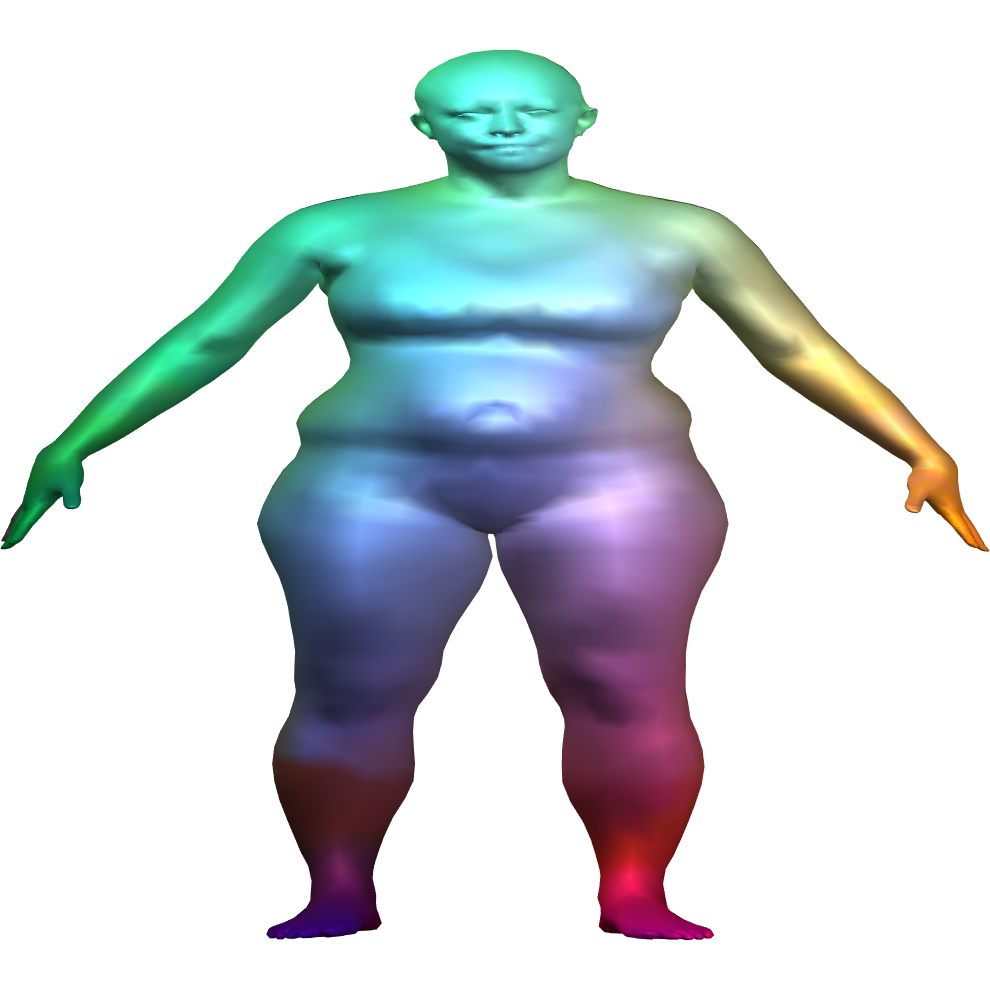} \\ 
\end{tabular}
\caption{Visualizing correspondence for varying number of descriptors for different algorithms. Functional Map: \cite{ovsjanikov2012functional}, Diagonal refers to \cite{nogneng2017informative} and Kernel Functional Map \cite{wang2018kernel} }.  
\label{fig:desc_map}
\end{table*} 
\clearpage
{\small
\bibliographystyle{ieee}
\bibliography{egpaper_for_review}
}
\end{document}